\documentclass{article}
\usepackage{icml2003}
\usepackage{epsf}
\usepackage{mlapa}
\usepackage{psfig}
\usepackage{subfigure}
\usepackage{graphicx}
\usepackage{color,epsfig}
\usepackage{algorithmic}
\usepackage{tabularx}
\usepackage[part,boxed]{algorithm}
\usepackage{wrapfig}
\usepackage{boxit}

\newtheorem{theorem}{Theorem}

\newtheorem{proposition}[theorem]{Proposition}

\def\l({\left(}      \def\r){\right)}

\newenvironment{pproof}[1][Proof]{\textbf{#1} }{\ \rule{0.5em}{0.5em}}

\newcommand{\CCC}{{\mathcal C}}

\begin{document}

\twocolumn[ \icmltitle{Hierarchical Multiclass Decompositions with Application
to Authorship Determination}

\icmlauthor{Ran El-Yaniv}{rani@cs.technion.ac.il}
\icmladdress{Department of Computer Science,\\
            Technion - Israel Institute of Technology}
\icmlauthor{Noam Etzion-Rosenberg}{noam@babylon.com}
\icmladdress{Babylon Ltd.\\
10 Hataasia St., Or-Yehuda, Israel}
\vskip 0.3in

\vskip 0.3in
]

\begin{abstract}
This paper is mainly concerned with the question of how
to decompose multiclass classification
problems into binary  subproblems.
We extend known Jensen-Shannon bounds on the Bayes risk
of binary problems to hierarchical multiclass problems
and use these bounds to develop a heuristic procedure for
constructing hierarchical multiclass decomposition
for multinomials.
We test our method and compare it to the
well known ``all-pairs'' decomposition.
Our tests are performed using a new authorship determination
benchmark test of machine learning authors.
The new method
consistently outperforms the all-pairs decomposition when the
number of classes is small and breaks even on larger
multiclass problems.
Using both methods, the classification accuracy we achieve,
using an SVM over a feature set consisting of both high frequency single tokens and
high frequency token-pairs, appears to be exceptionally
high compared to known results in authorship determination.
\end{abstract}

\section{Introduction}

In this paper we consider the problem of decomposing multiclass
classification problems into binary ones. While binary
classification is quite well explored, the question of multiclass
classification is still rather open and recently attracted
considerable attention of both machine learning theorists and
practitioners. A number of general decomposition schemes have
emerged, including `error-correcting output coding'
\cite{SejnowskiR87,DietterichB95}, the
more general
`probabilistic embedding' \cite{DekelS02} and
`constraint classification'
\cite{HarpeledRZ02}.
Nevertheless, practitioners are still mainly using the
infamous `one-vs-rest'
decomposition whereby an individual binary ``soft'' (or
confidence-rated) classifier is
trained to distinguish between each class and the union of the
other classes and then, for classifying an unseen
instance, all classifiers are applied and the winner classifier,
with the largest confidence for one of the classes, determines
the classification.
Another less commonly known method is the so called `all-pairs'
(or `one-vs-one') decomposition proposed by \cite{Friedman96}. In this method we train one
binary classifier for each pair of classes. To classify a new instance
we run a majority vote among all binary classifiers.
The nice property of the ``all-pairs'' method is that it generates
the easiest and most natural binary problems of all known methods.
The weakness of this method is that there may be
irrelevant binary classifiers which participate in the vote.
A number of papers provide evidences that `all-pairs' decompositions
are powerful and efficient and in particular, they outperform the `one-vs-rest'
method; see e.g. \cite{Furnkranz02}.


For the most part, known decomposition methods including
all those mentioned above are ``flat''.
In this paper we focus on \emph{hierarchical} decompositions.
The incentive to decompose a multiclass problem as
a hierarchy is natural and
can have at the outset general advantages which are both
statistical and computational.
Considering a multiclass problem with $k$ classes,
the idea is to learn a full binary tree\footnote{In a full binary tree
each node is either a leaf or has two children.}
of classes, where each node is associated
with a subset of the $k$ classes as follows: Each of the $k$
leaves is associated with a distinct class, and each internal node
is associated with the union of the class subsets of its right and left
children. Each such tree defines a hierarchical partition of the set of
classes and the idea is to train a binary classifier for each
internal node so as to discriminate between
the class subset of the right child and the class subset of the left child.
Note that in a full binary tree with $k$ leaves there are $k-1$
internal nodes.

Once these tree classifiers are trained, the classification
or ``decoding'' of an
new instance can be done using various approaches. One natural decoding
method would be to use the tree in a decision-tree fashion:
Start with the binary classifier at the root and let this classifier
determine either its right or left child, and this way follow
a path to a leaf and assign the class associated with this leaf.
This approach is particularly convenient when using
\emph{hard} binary classifiers giving labeles in $\{\pm 1\}$.
When using ``soft'' (confidence-rated) and in particular
probabilistic classifiers, giving confidence rates in $[0,1]$,
a natural decoding method would be to calculate an estimate
for the probability of following the path from the root
to each leaf and then
use a ``winner-takes-all'' approach, which selects the path with
the highest probability.

Besides computational efficiency,
the success of any multiclass decomposition scheme
depends on (at least) two interrelated factors. The first factor is the
statistical ``hardness'' of each
of the individual binary classification problems.
The second factor is the statistical robustness of the aggregation
(or ``decoding'') method.
The most fundamental measure for the hardness of a
classification problem is its \emph{Bayes error}.
We attempt to use the Bayes error of the resulting decomposition
and aim to hierarchically decompose the multiclass problem
so as to construct statistically ``easy'' collection of binary problems.

Determining the Bayes error of a classification problem based on the data
(and without
knowledge of the underlying distributions) is a hard problem,
without any restrictions \cite{LowerAntosDG99}.
In this paper we restrict ourselves to settings where
the underlying distributions can be faithfully modelled as \emph{multinomials}.
Potential application areas are classification of
natural language, biological sequences etc.
We can therefore in principle conveniently rely on studies,
which offer efficient and reliable
density estimation for multinomials
\cite{ristad95natural,friedman99efficient,llester00convergence,griths-using}.
As a first approximation, throughout this paper we make the assumption that we hold
``ideal'' data smaples and simply rely on maximum likelihood estimators that
count occurrences.

But even if the underlying distributions are known, a faithful estimation
of the Bayes error is computationally difficult.
We rely on known information theoretic bounds on the Bayes error, which
can be efficiently computed.
In particular, we use Bayes error bounds in terms of the Jansen-Shannon
divergence \cite{Lin91} and we derive upper and lower bounds
on the inherent classification difficulty of hierarchical
multiclass decompositions. Our bounds, which are tight
in the worst case, can be used as
optimality measures for such decompositions.
Unfortunatelly, the translation
of our bounds
into provably efficient algorithms to search for
high quality decompositions
appear at the moment computationally
difficult.
Therefore, we use a simple and efficient greedy
heuristic, which is able to generate reasonable
decompositions.

We provide initial empirical evaluation of our methods
and test them on multiclass problems of varying sizes
in the application area
of `authorship determination'. Our hierarchical decompositions
consistently improve on the `all-pairs' method when the number of classes
are small but do not outperform all-pairs with larger number of classes.
The authorship determination set of problems we consider is taken
from a new benchmark collection consisting of machine learning authors.
The absolute accuracy results we obtain are particularly high compared
to standard results in this area.

\section{Preliminaries: Bounds on the Bayes Error and the
Jensen-Shannon Divergence}

Consider a standard binary classification problem of classifying an observation
given by the random variable
$X$ into one of two classes $C_1$ and $C_2$. Let
$\pi_1$ and $\pi_2$ denote the priors on these two classes,
$\pi_1 + \pi_2=1$ with $\pi_i  \geq 0$.
Let $p_i(x) = p(X=x|C_i)$, $i=1,2$, be the class-conditional probabilities.
If $X=x$ is observed then by Bayes rule the posterior probability of $C_i$ is
$p(C_i|x) = \frac{\pi_i p_i(x)}{\pi_1 p_1(x) + \pi_2 p_2(x)}$.
If all probabilities are known we can achieve the Bayes error by choosing the
class with the larger posterior probability. Thus, the smallest error probability
is
$$
p(error|x) = \frac{ \min\{ \pi_1 p_1(x), \pi_2 p_2(x)\} }
                  {\pi_1 p_1(x)+\pi_2 p_2(x)},
$$
and the Bayes error is given by
$p_{Bayes} = p(error) = \int_x p(x) p(error|x)dx = E_x [\min\{ \pi_1 p_1(x), \pi_2 p_2(x)\}]$.

The Bayes error quantifies the inherent difficulty of the classification problem
at hand (given the entire probabilistic characterization of the problem)
without any considerations of inductive approximation based on finite
samples. In this paper we attempt to decompose multi-class problems
into hierarchically ordered
collections of binary problems so as to minimize the Bayes error of
the entire construction.

\subsection{The Jensen-Shannon (JS) Divergence}
Let $P_1$ and $P_2$ be two distributions over some finite set $X$,
and let $\pi = (\pi_1,\pi_2)$ be their priors.
Then, the Jensen-Shannon (JS) divergence \cite{Lin91} of $P_1$ and and $P_2$
with respect to the prior $\pi$
is
\begin{equation}
\label{eq:js_entropy}
JS_{\pi}(P_1,P_2) = H(\pi_1 P1 + \pi_2 P_2) - \pi_1 H(P_1) - \pi_2(P_2),
\end{equation}
where $H(\cdot)$ is the Shannon entropy.
It can be shown that $JS_{\pi}(P_1,P_2)$ is non-negative, symmetric,
bounded (by $H(\pi)$) and it equals zero if and only if
$P_1 \equiv P_2$.
According to \cite{Lin91} the JS-divergence was first introduced by \cite{WongY85}
as a dissimilarity measure for random graphs.
Setting $M_{\pi} = \pi_1 P1 + \pi_2 P_2$
it is easy to see \cite{ElYanivFT97} that
\begin{equation}
\label{eq:js_dkl}
JS(P_1,P_2) = \pi_1 D_{KL}(P_1 || M_{\pi}) + \pi_2 D_{KL}(P_2 || M_{\pi}),
\end{equation}
where $D_{KL}(\cdot || \cdot)$ is the Kullback-Leibler divergence
\cite{CoverT91}.
The average distribution $M_{\pi}$ is called the \emph{mutual source}
of $P_1$ and $P_2$ \cite{ElYanivFT97} and it can be easily shown
that
\begin{equation}
\label{eq:mutual}
M_{\pi} = \arg \min_{Q} \pi_1 D_{KL}(P_1 || Q) + \pi_2 D_{KL}(P_2 || Q).
\end{equation}
That, is the mutual source of $P_1$ and $P_2$ is the closest
to both of them simultaneously in terms of the KL-divergence.
Like the KL-divergence the JS-divergence has a
number of important roles in statistics and pattern recognition.
In particular, the JS-divergence, compared against a threshold is
an optimal statistical test in the Neyman-Pearson sense
 \cite{Gutman89} for the \emph{two-sample problem} \cite{Lehmann59}.


\subsection{Jensen-Shannon Bounds on the Bayes Error}
Lower and upper bounds on the binary Bayes error are
given by \cite{Lin91}. Again, let $\pi = (\pi_1,\pi_2)$ be the priors and $p_1, p_2$, the
class conditionals, as defined above. Let $p(error)$ be the Bayes error.
Set $J = H(\pi) - JS_{\pi}(p_1,p_2)$ with $H(\pi)$ denoting the binary entropy.
\begin{theorem}[Lin]
\label{thm:lin2}
\begin{equation}
\label{eq:lin2}
\frac{1}{4}J^2 \leq p(error) \leq \frac{1}{2} J
\end{equation}
\end{theorem}

These bounds are generalized to $k$ classes in a straightforward
manner. Considering a multiclass problem with $k$ classes
and class-conditionals $p_1,\ldots,p_k$ and priors
$\pi = (\pi_1,\ldots,\pi_k)$, the Bayes error is given by
$$
p({error}_k) = \int_x p(x)(1 - \max\{p(C_1|x),\ldots,p(C_k|x)\})dx.
$$
Now setting $J_k = H(\pi) - JS_{\pi}(p_1,\ldots,p_k)$ we have
\begin{theorem}[Lin]
\label{thm:lin_multi}
\begin{equation}
\label{eq:lin_bound_general}
\frac{1}{4(k-1)}J_k^2 \leq p({error}_k) \leq \frac{1}{2}J.
\end{equation}
\end{theorem}
Given the true class-conditional, these JS bounds on the
Bayes error can be efficiently computed using
either (\ref{eq:js_entropy}) or (\ref{eq:js_dkl}) (or their generalized
forms).

\section{Bounds on the Bayes Error of Hierarchical Decompositions}
\label{sec:bounds}

In this section we provide bounds on the Bayes
error of hierarchical decompositions. The
bounds are obtained using a straightforward
application of the binary bounds of Theorem~\ref{thm:lin2}.
We begin with a more formal description of hierarchical decompositions.

Consider a multi-class problem with $k$ classes $\CCC = C_1,\ldots,C_k$,
and let $T=(V,E)$ be any full binary tree with $k$ leaves, one for each class.
For each node $v \in V$ we map a label set $\ell(v) \subseteq \CCC$ which is defined
as follows.
Each leaf $v$ (of the $k$ leaves) is mapped to a unique
class (among the $k$ classes).
If $v$ is an internal node whose left and right children are $v_{L}$ and $v_R$, respectively,
then $\ell(v) = \ell(v_L) \cup \ell(v_R)$.
Given the tree $T$ and the mapping $\ell$ we decompose the multi-class problem by
constructing a binary classifier $h_v$ for each internal node $v$ of $T$ such that
$h_v$ is trained to discriminate between classes in
$\ell(v_L)$ and classes in $\ell(v_R)$.
In the case
of hard classifiers $h_v(x) \in \{ \pm 1\}$ and we identify `$-1$' with `$L$'
and `$+1$' with `$R$'. In the case of soft classifiers, $h_v(x)\in [0,1]$
and we identify 0 with `$L$' and 1 with `$R$'.
Since there are $k$ leaves there are exactly $k-1$
binary classifiers in the tree. The training set of each
classifier is naturally determined by the  mapping $\ell$.

Given a sample $x$ whose label (in $\CCC$) is unknown, one can think of a number of
``decoding'' schemes that combine the individual binary classifiers.
When considering hard binary classifiers a natural choice to aggregate the
binary decisions is to start from the root $r$ and apply its associated
classifier $h_r$. If $h_r(x) = -1$ we go to $r_L$ and otherwise we go to $r_R$, etc.
This way we continue until we reach a leaf and predict for $x$ this leaf's associated
(unique) class. In the case of soft binary classifiers
a natural decomposition
would be to consider for each leaf $v$
the path from the root to $v$, and multiply the probability estimates
along this path.
Then the leaf with the largest probability will assign a label to $x$.

There is a huge number of possible hierarchical decompositions already for moderate
values of $k$.
We note that a known decomposition scheme which is captured by such hierarchical
constructions is the decision list multiclass decomposition approach
(referred to as ``ordered one-against-all class binarization'' in
\cite{Furnkranz02}).

Consider a $k$-way multiclass problem with class conditionals
$P_i(x) = P(x|C_i)$ and priors $\pi_1,\ldots,\pi_k$.
Suppose we are given a decomposition structure $(T,\ell)$ for $k$ classes consisting
of the tree $T$ and the class mapping $\ell$.
Each internal node $v$ of $T$ corresponds
to one binary classification problem. The original multiclass
problem naturally induces class conditional probabilities and priors
for the binary problem at $v$ and we
denote these conditionals by $p_v(x|v_L)$ and $p_v(x|v_R)$ and the prior by
$\pi(v)$.
For example, denoting the root of $T$ by $r$, we have
$$
p_r(r_L|x) = \sum_{C_i \in \ell(r_L)}p(C_i|x),
$$
with $p_r(x|r_L) = p_r(r_L|x)p(x)/\pi(r_L)$ by Bayes rule and
$\pi(L) = \sum_{C_i \in \ell(r_L)} \pi_i$.
Let $p_v(error)$ be the Bayes error of this problem and denote
the Bayes error of the entire tree by $p_T(error)$.
\begin{proposition}
\label{thm:2_upper}
For each internal node $v$ of $T$ let
$q(v) = (1- \frac{1}{2}J(v))$
where
$$
J(v) = H\left[\pi(v)\right] - JS_{\pi(v)}\left[p_v(x|v_L),p_v(x|v_R)\right].
$$
Then
$$
p_T(error) \leq 1 - Q(T),
$$
where
\begin{equation}
\label{eq:QT}
Q(T) = q(r)\left[Q(T_L)+Q(T_R)\right]
\end{equation}
and for a leaf $v$, $Q(v) = 1$.
\end{proposition}

\begin{pproof}
For each class $j$, $j=1,\ldots,k$
let $v^j_1,v^j_1,\ldots,v^j_{n_j}$ be the path from the root to the leaf corresponding
to class $j$, where $v^j_1$ is the root of $T$ and
$v^j_{n_j}$ is the leaf. This path consists of $n_j - 1$ binary problems.
The probability of following this path and reaching the leaf $v_{n_j}$ is
$$
P[\textrm{reaching $v^j_{n_j}$}] = \prod_{i=1}^{n_j -1} (1- p_{v^j_i}(error)).
$$
Thus, the overall average error probability
$P_T(error)$ for the entire structure $(T,\ell)$ is
\begin{eqnarray*}
P_T(error) &= \sum_{j=1}^k \pi_j (1 - P[\textrm{reaching $v^j_{n_j}$}]) \\
&= 1 - \sum_{j=1}^k \prod_{i=1}^{n_j -1} (1- p_{v^j_i}(error)).
\end{eqnarray*}
Using the JS (upper) bound from Equation~(\ref{eq:lin2})
on the individual binary problems in $T$ we have
\begin{equation}
\label{eq:tree_bayes_bound}
P_T(error) \leq 1 - \sum_{j=1}^k \prod_{i=1}^{n_j -1} (1- \frac{1}{2}J(v^j_i)),
\end{equation}
where for $v=v^j_i$ $J(v) = H(\pi(v)) - JS_{\pi(v)}(p_v(x|v_L),p_v(x|v_R))$.
Rearranging terms it is not hard to see that
$$
Q(T) = \sum_{j=1}^k \prod_{i=1}^{n_j -1} (1- \frac{1}{2}J(v^j_i))
$$
\end{pproof}

The same derivation now using the
JS lower bound of Equation~(\ref{eq:lin2}) yields:

\begin{proposition}
\label{thm:2_lower}
For each internal node $v$ of $T$ let
$q'(v) = (1- \frac{1}{4}J'(v))$
where
$$
J'(v) = \left(H\left[\pi(v)\right] - JS_{\pi(v)}\left[p_v(x|v_L),p_v(x|v_R)\right]\right)^2.
$$
Then
$$
p_T(error) \geq 1 - Q'(T),
$$
where
$$
Q'(T) = q'(r)\left[Q'(T_L)+Q'(T_R)\right]
$$
and for a leaf $v$, $Q(v) = 1$.
\end{proposition}

\section{A Heuristic Procedure for Agglomerative Tree Constructions}

The recurrences of Propositions~\ref{thm:2_upper} and \ref{thm:2_lower} provide
the means for efficient calculations of upper and lower bounds on the
multiclss Bayes error of any tree decomposition
given the class conditional probabilities
of the leaves.
Our goal is to construct a
full binary $T$ whose Bayes error is minimal.
A natural approach would be to consider trees whose Bayes error
upper bound are minimal.
This corresponds to maximizing $Q(T)$ (\ref{eq:QT}) over all trees $T$.
There are two obstacles for achieving this goal.
The statistical obstacle is that
the true class conditional distributions of internal nodes are
not available to us. The computational obstacle is that
the number of possible trees is huge.\footnote{The
number of unlabeled full binary trees with $k$ leaves
is the Catalan number $C_{k-1} = \frac{1}{k}{{2k-2}\choose{k-1}}$.
The number of labeled trees (not counting isomorphic trees) is $O(2^k k!)$.}
Handling the first obstacle in the general case using density estimation technics appears
to be counterproductive as density estimation is considered harder than
classification. But we can restrict ourselves to parametric models such
as multinomials where estimation of the class conditional probabilities
can be achieved reliably and efficiently; see e.g.
\cite{ristad95natural,friedman99efficient,llester00convergence,griths-using}.
In the present work we ignore the discrepancy that will
appear in our Bayes error bounds (even in the case of multinomials)
and rely on simple maximum likelihood estimates
of the class-conditionals.

To handle the maximization of $Q(T)$ we use the following
agglomerative randomized heuristic procedure.
We start with a forest of all $k$ leaves, corresponding
to the $k$ classes. Our estimates for the prior of these
classes $\pi_j$, $j=1,\ldots,k$, are obtained from the data.
We perform $k-1$ agglomerative mergers as follows.
On step $i$, $i=1,\ldots,k-1$ we have a forest $F_i$ containing
$N_i = k-i+1$ trees, $T_1,\ldots,T_{N_i}$. Each of these trees $T$
has an associated class-conditional probability $P_T(x)$ (which
is again estimated from the data), and a weight
$w(T)$ that equals the sum of priors of its leaves.
For each pair of trees $T_i$ and $T_j$ we compute
their JS-divergence $JS(i,j) = JS_{\pi(i,j)}(P_{T_i}(x),P_{T_j}(x))$
where $\pi(i,j) = \left(w(T_i)/(w(T_i)+w(T_j)),w(T_j)/(w(T_i)+w(T_j))\right)$.
For each possible merger (between $i$ and $j$) we assign
the probability $p(i,j)$ proportional to $2^{-JS(i,j)}$.
This way large JS values are assigned to smaller probabilities
and vice versa.\footnote{Using
a Bayesian argument it can
be shown \cite{ElYanivFT97} that if
$X$ and $Y$ are samples with types (empirical probability) $P_{T_i}$ and $P_{T_j}$,
respectively, then
$2^{-JS(i,j)}$ is proportional
to the probability
that $X$ and $Y$ emerged from
the same distribution.}
We then randomly choose one merger according to these probabilities.
The newly merged tree $T_{ij}$ is assigned the mutual source of
$T_i$ and $T_j$ as its class-conditional (see Equation~(\ref{eq:mutual}))
and its weight is $w(T_i)+w(T_j)$.
In all the experiments described below,
to obtain a multiclass decomposition we ran this randomized procedure
10 times and chose the tree $T$ that maximized $Q(T)$.
The chosen tree $T$ then determines the hierarchical
decomposition, as described in Section~\ref{sec:bounds}.
Note that the above procedure does not directly maximize
$Q(T)$. The routine simply attempts to find trees whose higher internal nodes
are ``well-separated''. Such trees will have low Bayes error and our
formal indication for that will be that $Q(T)$ will be large.
Thus, currently we can only use our bounds as a means to \emph{verify} that a hierarchical
decomposition is good, or to \emph{compare} between two decompositions.

\section{The Machine Learning Authors Dataset}
\label{sec:MLA}

In our experiments (Section~\ref{sec:experiments})
we used a new benchmark dataset for testing authorship
determination algorithms. This dataset contains a
collection of singly-authored scientific research papers.
The scientific affiliation of all authors is machine learning,
statistical pattern recognition
and related application areas. After this dataset was automatically collected
from the web using a focused crawler guided by a compiled list of machine learning
researchers, it was manually checked to see that all papers are indeed by
single authors.
This \emph{Machine Learning Authors (MLA)} dataset. contains
articles by more than 400 authors with each author having at least
one singly-authored paper.\footnote{The MLA dataset will soon be
publicly available at http://www.cs.technion.ac.il/$\sim$rani/authorship.}
For the present study we extracted from the MLA collection a
subset that was prepared as follows. The raw papers (given in
either PS or PDF formats) were first translated to ascii and then
each paper was parsed into \emph{tokens}. A token is either a word
(a sequence of alpha numeric characters ending with one of the
space characters or a punctuation) or a punctuation
symbol.\footnote{ We considered as tokens the following
punctuations: .;,:?!'()"-/$\backslash$.} To enhance uniformity and
experimental control we segmented each paper into chunks of
\emph{paragraphs} where a paragraph contains 1000
tokens.\footnote{Last paragraphs of length $< 500$ tokens were
combined with second-last paragraphs. This way, paragraphs lengths
vary in $[500,1499]$ but a large majority of the paragraphs are of
exactly 1000 tokens.} To eliminate topical information we
projected all documents on the most frequent 5000 tokens.
Appearing among these tokens are almost all of the most frequent
\emph{function words} in English, which bare no topical content
but are known to provide highly discriminative information for
authorship determination \cite{MostellerW64,Burrows87}. For
example, on Figure~\ref{fig:projected_text} we see a projected
excerpt from the paper \cite{mitchell99machine} as well as its
source containing all the tokens. Clearly there are non-function
words (like `data'), which remained in the projected excerpt.
Nevertheless, since all the authors in the dataset write about
machine learning related issues, such words do not contain much
topical content.


We selected from MLA only the authors who have more than
30 paragraphs in the dataset. The result is a set of exactly 100 authors
and in the rest of the paper we call the resulting set
the MLA-100 dataset.

\begin{figure}[htb!]
\fbox{\begin{minipage}{0.9\columnwidth}
\centerline{\bf Projected Text}
\smallskip
{\footnotesize
Over the many have to of data their ,,their ,,and
their ..At the same time,,,and in many nd complex ,,such as the of data that in ..
The of data the of how best to use this data to general and to ..Data ::using data
to and ..The of in data follows from the of several :
}
\end{minipage}}

\fbox{\begin{minipage}{0.9\columnwidth}
\centerline{\bf Original Text}
\smallskip
{\footnotesize
Over the past decade many organizations have begun to routinely capture
huge volumes of historical data describing their operations, their products,
and their customers. At the same time, scientists and engineers in many
fields find themselves capturing increasingly complex experimental datasets,
such as the gigabytes of functional MRI data that describe brain activity
in humans. The field of data mining addresses the question of how best to
use this historical data to discover general regularities and to improve future
decisions.
Data Mining: using historical data to discover regularities and
improve future decisions.
The rapid growth of interest in data mining follows from the confluence
of several recent trends:
}
\end{minipage}}
\caption{
An excerpt from
the paper ``Machine Learning and Data Mining'' \cite{mitchell99machine}.
Top: A projection of the text over the high frequency tokens; Bottom: The original text.
Excerpt is taken from
the paper Machine Learning and Data Mining \cite{mitchell99machine}.
\label{fig:projected_text}}
\end{figure}

\section{Experiments}
\label{sec:experiments}
Here we describe our initial empirical studies of the proposed multiclass
decomposition procedure.
We compare our method with
the ``all-pairs' decomposition.
Taking the MLA-100 dataset (see Section~\ref{sec:MLA}) we generated a
a progressively increasing random subset as follows. From the MLA-100
we randomly chose
3 authors, then added another author, chosen randomly and uniformly
from the remaining authors, etc. This way we generated increasing sets of authors
in the range of 3-100. So far we have experimented with multiclass subsets with
$k=3-20,50$ and $100$.
In all the experiments we used an SVM with an RBF kernel.
The SVM parameters where chosen using cross-validation.
The reported results are averages of 3-fold cross-validation.

The features generated for our authorship determination
problems contained in all cases the top 5000 single tokens
(see Section~\ref{sec:MLA} for the token definition) as well as the
following ``high order pairs''. After projecting the documents over the high frequency
single tokens we took all bigrams. For instance, considering the projected
text in Figure~\ref{fig:projected_text}, the token pair
`to'$+$`of' appearing in the first line of the projected text (top)
is one of our features. Notice that in the original text this pair
of words appears 5 words apart. This way our representation
captures high order pairwise statistics of the tokens.
Moreover, since we restrict ourselves to the most frequent
tokens in the text these pairs of token do not suffer too much from
the typical statistical sparsness which is usually experienced
when considering $n$-grams in text categorization and language models.

\begin{figure}[htb!]
  \begin{center}
      \epsfig{file=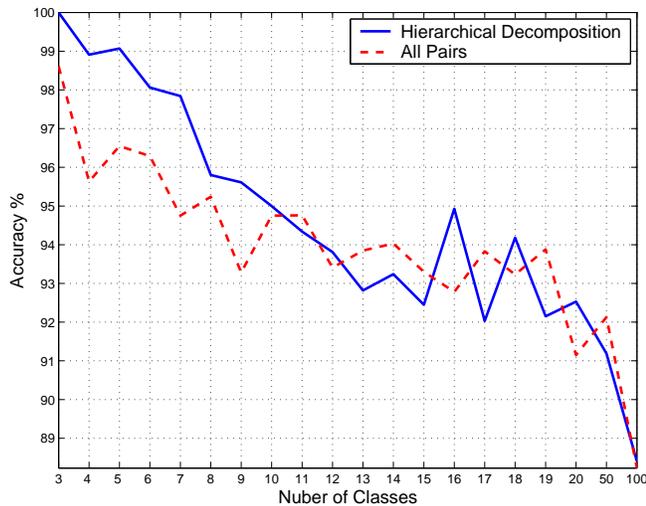,width=3.4in}
    \caption{The performance of hierarchical multiclass decompositions
    and `all-pairs' decompositions on 20 authorship determination problems
    with varying number of classes.
\label{fig:performance}}
\end{center}
\end{figure}

Accuracy results for both ``all-pairs'' and our hierarchical
decomposition procedure appear in Figure~\ref{fig:performance}.
The first observation is that the absolute values
of these classification results are rather high compared
to typical figures reported in authorship determination.
For example, \cite{StamatatosFK} report on accuracy around 70\% for
determining between 10 authors of newspaper articles. Such figures
(i.e. number of authors and around 60\%-80\% accuracy) appear to be common
in this field.
The closest results in both size and accuracy we have found are of
\cite{Rao00}, who distinguish between 117 newsgroup authors
with accuracy 58.8\% and between 84 authors with accuracy 80.9\%.
Still, this is far from he 91\% accuracy
we obtain for 50 authors and 88\% accuracy for 100 authors.

The consistent advantage of hierarchical decompositions over all-pairs is evident for
small number of classes. However, for over 10 classes, there is no significant
difference between the methods.
Interestingly, the best hierarchical constructs our method generated
(in terms of the $Q(T)$) were completely skewed. It is not clear to us
at this stage whether
this is an artifact of our Bayes error bound or a weakness of our heuristic
procedure.


\section{Concluding Remarks}
This paper presents a new approach for hierarchical multiclass decomposition
of multinomials.
A similar hierarchical approach can be attempted with nonparameteric models.
For instance using any nonparametric probabilistic binary discriminator
one can attempt to heuristically estimate the hardness of the involved
binary problems using empirical error rates and design reasonable hierarchical
decompositions.
However, a major difficulty in this approach
is the computational burden.

When considering the main inherent deficiency of all-pairs decompositions
it appears that this deficiency should disappear or at least soften when
the number of classes increases. The reason is that with large number of
classes, the noisy votings of irrelevant classifiers will tend to cancel
out and the power of the relevant classifiers will then increase.
We therefore speculate that it would be very hard to consistently
beat all-pairs decompositions
with very large number of classes.
Nevertheless,a desirable property of a decomposition
scheme is \emph{scalability}, which allows for efficient handling
of large number of classes (and datasets).
For example, one can hypothesize useful authorship determination
applications, which need to determine between thousands or even millions
of authors. While balanced hierarchical decomposition will be able
to scale up to these dimensions, the $O(k^2)$ complexity of the
all-pairs method would probably start to form a computational bottleneck.

\bibliography{active2}
\bibliographystyle{mlapa}

\end{document}